\documentclass[runningheads]{llncs}
\usepackage{graphicx}
\usepackage{amsmath,amssymb} 
\usepackage{color}

\usepackage{amsfonts}
\usepackage{array}
\usepackage{booktabs}
\usepackage{bm}
\usepackage{enumitem}
\usepackage{makecell}
\usepackage{multirow}
\usepackage{subcaption}
\usepackage{tabularx}
\usepackage{ulem}
\usepackage{xcolor}
\usepackage{xpatch}
\usepackage{hyperref}

\def\tablefontsize{\footnotesize}
\DeclareMathOperator{\binarize}{Binarize}
\newcommand\Mark[1]{\textsuperscript#1}

\begin{document}
\pagestyle{headings}
\mainmatter


\title{SpotPatch: Parameter-Efficient Transfer Learning for Mobile Object Detection} 
\titlerunning{SpotPatch: Parameter-Efficient Transfer Learning}
%
%
\authorrunning{K. Ye et al.}
%
\author{Keren Ye\thanks{Work partially done during an internship at Google.}\Mark{1}
\and Adriana Kovashka\Mark{1}
\and Mark Sandler\Mark{2}
\and Menglong Zhu\thanks{Work done at Google.}\Mark{3}
\and \\Andrew Howard\Mark{2} 
\and Marco Fornoni\Mark{2}}

\institute{
\Mark{1}University of Pittsburgh \hspace{0.4cm}
\Mark{2}Google Research \hspace{0.4cm} 
\Mark{3}DJI Technology LLC \\
\texttt{\small{\{yekeren, kovashka\}@cs.pitt.edu \hspace{0.4cm} sandler@google.com \\ mzhu.upenn@gmail.com \hspace{0.4cm} \{howarda, fornoni\}@google.com}}
}


\maketitle

\begin{abstract}
Deep learning based object detectors are commonly deployed on mobile devices to solve a variety of tasks. For maximum accuracy, each detector is usually trained to solve one single specific task, and comes with a completely independent set of parameters. While this guarantees high performance, it is also highly inefficient, as each model has to be separately downloaded and stored. In this paper we address the question: can task-specific detectors be trained and represented as a shared set of weights, plus a very small set of additional weights for each task? The main contributions of this paper are the following: 1) we perform the first systematic study of parameter-efficient transfer learning techniques for object detection problems; 2) we propose a technique to learn a model patch with a size that is dependent on the difficulty of the task to be learned, and validate our approach on 10 different object detection tasks. Our approach achieves similar accuracy as previously proposed approaches, while being significantly more compact.
\end{abstract}
\section{Introduction}
\label{sec:intro}

Mobile object detection models are fundamental building blocks for daily-used mobile applications. For example, face detectors are used for locking/unlocking the latest generation phones and for building social apps such as Snapchat.
In the early years, most computer vision models were deployed on servers, which meant that images had to be sent from the device to the server and that users had to wait for the server responses. This process was sensitive to network outages, provided on-device latency that was often not tolerable, and burdened the server clusters with high loads of client requests.

With the advance of mobile hardware technology, on-device computation became more and more affordable. Meanwhile, advances in neural network architectures made model inference increasingly more efficient.
On the one hand, MobileNets~\cite{Howard_2019_ICCV,howard2017mobilenets,Sandler_2018_CVPR} optimize the network architecture by decomposing convolutions into more efficient operations. Such designs provide general and compact backbones for mobile inference. On the other hand, one-stage detection architectures such as SSD~\cite{liu2016ssd} and Yolo~\cite{redmon2016you} provide mobile-friendly detection heads.

\begin{figure}[t]
    \centering
    \includegraphics[width=0.6\linewidth]{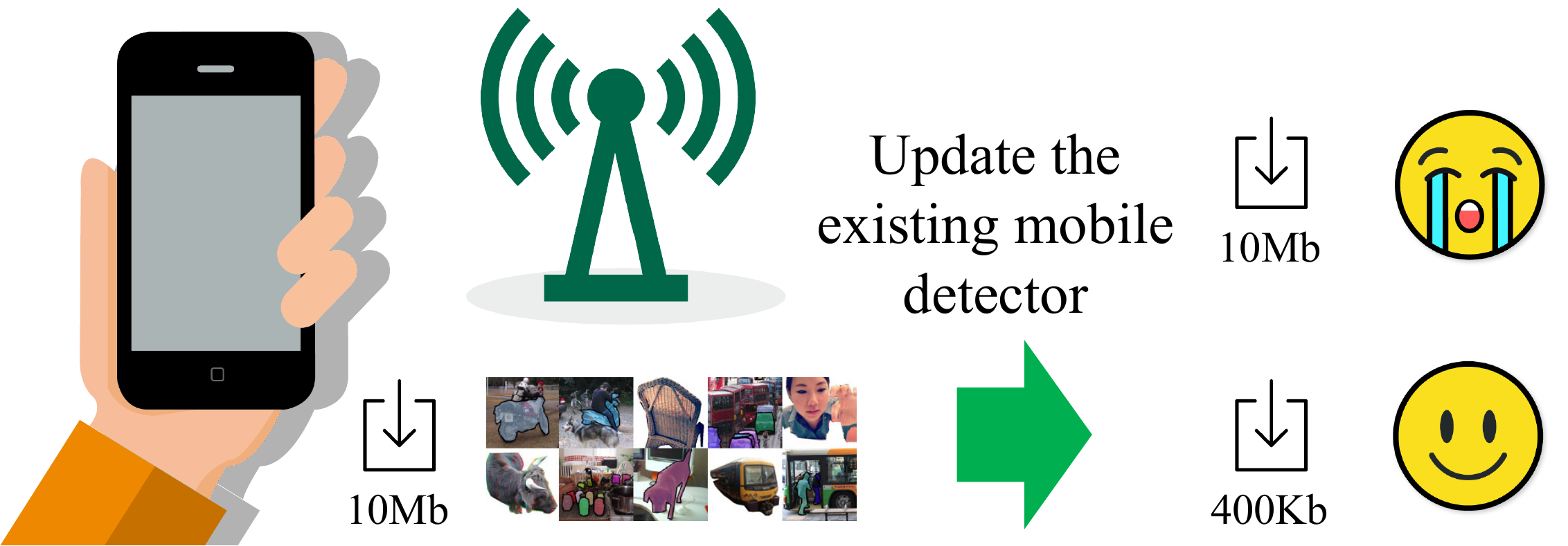}
    \caption{\textbf{The challenge of updating a mobile object detector.} Suppose a general-purpose object detector is already deployed on-device. In a naive setting, adding support for detecting new entities would require downloading a completely separate model, with large network costs. Our goal is to reduce the network costs by \textit{``patching''} the existing model to also support the new entities.}
    \label{fig:concept}
\end{figure}

Due to the combination of the above advancements, object detection models are now massively being moved from server-side to on-device. While this constitutes great progress, it also brings new challenges (see Fig.~\ref{fig:concept}). Specifically, multiple isolated models are often downloaded to perform related tasks.
Suppose that a well-performing mobile model is downloaded for general-purpose object detection, with 10MB data traffic costs. Suppose then that the user requests an additional functionality that requires detecting new entities, like faces or barcodes: this will naively require to download a new model, with an extra 10MB data cost. Each time a new task will need to be supported, the user and the network operator will incur an additional 10MB data cost. The question is then: can we instead ``patch'' the previously downloaded general-purpose object detector to solve also the new tasks?
If so, how can we minimize the size of the ``patch'' while maintaining high accuracy on the new task? To answer these questions, we studied two experimental scenarios that mimic two practical use-cases for ``patching'' mobile detection models:

\begin{enumerate}[nolistsep,noitemsep]
    \item Adapting an object detector to solve a new task.
    \item Updating an existing model, whenever additional training data is available.
\end{enumerate}

\begin{figure*}[t]
    \centering
    \includegraphics[width=1.0\linewidth]{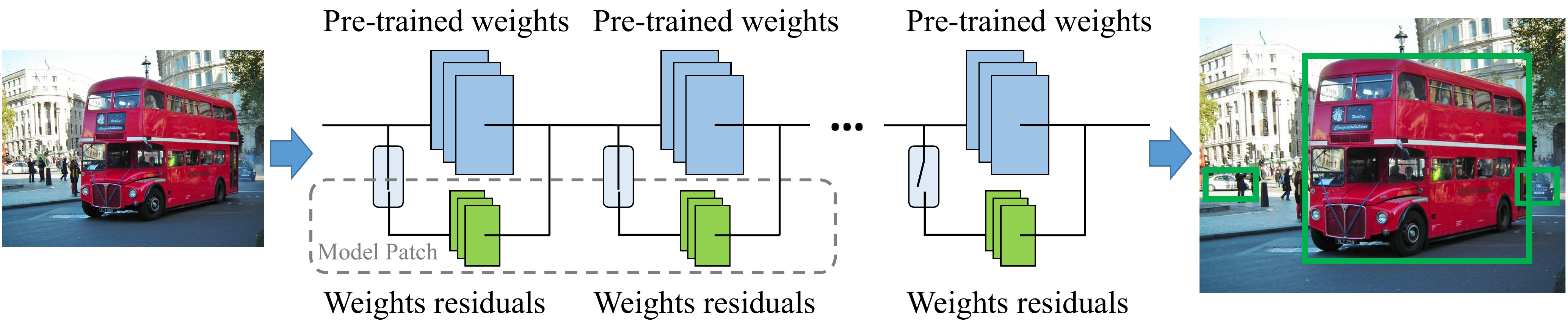}
    \caption{\textbf{SpotPatch: demonstrating the gating mechanism.} During training, our model optimizes both the detection performance and the number of patched layers by opening, or closing per-layer gates. During deployment, only the \textit{opened routes} constitute the model patch. We use a 1-bit representation for the weights residuals, significantly reducing the patch footprint.}
    \label{fig:gating}
\end{figure*}

To learn the model patch, we propose an approach simultaneously optimizing for accuracy and footprint (see Fig.~\ref{fig:gating} for an overview): 1) for each layer, we minimize the size of the patch by using a 1-bit representation of the weight residuals; 2) we employ a gating mechanism to selectively patch only important layers, while reusing the original weights for the remaining ones.
We evaluate our problem on ten different object detection tasks, using an experimental setup similar to~\cite{rebuffi2017learning}, which we refer to as ``Detection Decathlon''. We also showcase our method's ability to efficiently update the model when new data becomes available. To the best of our knowledge, this is the first systematic study of parameter-efficient transfer learning techniques on object detection tasks.

\section{Related Work}
\label{sec:related}

The most relevant approaches for our work fall into three main categories:
(1) Model footprint reduction, aimed at reducing the number of trainable parameters, or the bit-size representation for each parameter.
(2) Dynamic routing, adapting the network architecture at training time, based on well-designed cost functions.
(3) Transfer learning and domain adaptation, ensuring representation transferability across tasks and domains.

\textbf{Reducing the model footprint} works involve various approaches that directly reduce the model size.
Instead of fine-tuning the whole network~\cite{Cui_2018_CVPR} or the last few layers~\cite{Oquab_2014_CVPR,Thomas_2016_CVPR,NIPS2014_5347}, \cite{Mudrakarta_2019_ICLR} proposed to learn a set of parameters dispersed throughout the network. They experimented with updating batch normalization and depthwise convolution kernels, resulting in small model patches providing high classification accuracy. Similar findings were also published in~\cite{Guo2019DepthwiseCI}.

Quantization is a technique commonly used in on-device models to reduce model size. Post-training quantization~\cite{post_training_quantization} can be applied to quantize a pre-trained floating-point precision model, while quantization-aware training \cite{baskin2018uniq,Jacob_2018_CVPR,Hubara_2017_JMLR,quantization_aware_training} ensures that the quantization effects are already modeled at training time.

Ideas inspired by the quantization literature are used in transfer and multi-task learning to reduce the footprint of the model for the target tasks. For example, in~\cite{Mallya_2018_ECCV}, learned binary masks are element-wise multiplied to the source kernels to obtain the convolutional kernels to be used on the target task. In~\cite{Mancini_2018_ECCVW}, the binary masks are augmented with floating-point scalars to produce an affine transformation of the kernel. 
Our method learns similar kernel masks,
but we go one step further by automatically selecting the subset of layers to patch.

\textbf{Dynamic routing} works adapt the network structure to optimize a pre-defined objective.
\cite{shazeer2017outrageously} trained a gating network selecting a sparse combination of experts based on input examples.
\cite{liu2018dynamic} created a model allowing selective execution. In their setting, given an input, only a subset of neurons is executed.
\cite{Veit_2018_ECCV} reduced the number of ResNet~\cite{He_2016_CVPR} layers by bypassing residual blocks using a gating mechanism.
\cite{Guo_2019_CVPR} built a dynamic routing network to choose between using either the pre-trained frozen blocks, or the re-trained blocks.

Our approach differs from all these studies in that: (1) our dynamic model architecture is conditioned on dataset rather than on input; (2) we optimize (reduce) the number of patched layers; and (3) one of the route types, in our design, is specialized to use binary weights to further reduce footprint.

\textbf{Transfer learning and domain adaptation} works study the ability of models to transfer across tasks, in terms of achieving optimal accuracy in novel situations. 
In transfer learning, the target label set might differ from the source task. In domain adaptation, the classes may be the same, but have inherently different distributions.
Some transfer and adaptation techniques minimize the discrepancy between tasks or domains in terms of the feature representation \cite{Carlucci_2019_CVPR,Chen_2018_CVPR,ganin2016domain,herath2019min,long2016unsupervised,sun2019meta,Tzeng_2017_CVPR,Vu_2019_ICCV,wang2019few,yin2019feature}. If the classes are the same, the deviation between classifier weights across domains can be minimized \cite{kovashka2013adaptation,yang2007adapting}.
One of the most established recent benchmarks for transfer of classification networks is the Visual Decathlon Challenge~\cite{rebuffi2017learning}, in which a decathlon-inspired scoring function is used to evaluate how well a single network can solve 10 different classification tasks.
In the object detection literature, \cite{Uijlings_2018_CVPR} proposes to use a source domain to generate boxes of different levels of class-specificity, to transfer to a target domain where only image-level labels are available. 
\textit{In contrast to optimizing accuracy despite limited data in the target domain,} as is often the objective of domain adaptation, in this work we are concerned with preserving high accuracy while \textit{minimizing the footprint} of the task-specific model patches.
\section{Approach}
\label{sec:approach}

To simplify notation, we assume a deep neural network $\mathcal{M}$ of depth $N$ is composed of a set of layers represented by their weights $\bm{\theta} = \{\mathbf{W}_1, \dots, \mathbf{W}_N\}$, plus an activation function $\Phi$. The transformation computed by the network is represented using Eq.~\ref{eq:deep_net}, where $\bm{x}$ is the input and $\bm{z}_i$ denotes the $i$-th hidden state.
\begin{equation} \begin{split}
\label{eq:deep_net}
    \left\{ \begin{aligned}
        &\bm{z}_i = \Phi(\mathbf{W}_i \bm{z}_{i-1}), ~~~~ \bm{z}_0 = \bm{x} \\
        &\mathcal{M}(\bm{x}) = \mathbf{W}_N \bm{z}_{N-1}
    \end{aligned} \right.
\end{split} \end{equation}

To adapt $\mathcal{M}$ to solve a new task, we seek a task-specific parameter $\bm{\theta}' = \{\mathbf{W}_1', \dots, \mathbf{W}_N'\}$ that optimizes the loss on the target dataset.
In addition, since we do not want to fully re-learn $\bm{\theta}'$, we look for a transformation with minimum cost (measured in terms of footprint) to convert the original $\bm{\theta}$ to $\bm{\theta}'$.
Assume the transformation function can be expressed as $\bm{\theta}'= f(\bm{\theta}, \bm{\gamma})$ where $\bm{\gamma}$ is an additional set of parameters for the new task, and $f$ is a function that combines the original parameters $\bm{\theta}$ with the parameter ``patch'' $\bm{\gamma}$ (as a very simple example, through addition). Our goal is to reduce the bit size of $\bm{\gamma}$.
In our experiments, we use relative footprint $\frac{\text{bitsize}(\bm{\gamma})}{\text{bitsize}(\bm{\theta})}$ and patch size $\text{bitsize}(\bm{\gamma})$ as footprint metrics.

We propose two approaches to compress the patch $\bm{\gamma}$,
 namely \textit{task-specific weight transform} and \textit{spot patching}.
The former is inspired by the early Adaptive-SVM approaches such as \cite{kovashka2013attribute,yang2007adapting}, their Deep Neural Network counterparts such as ~\cite{Mancini_2018_ECCVW}, as well as quantization methods~\cite{Jacob_2018_CVPR} and the low-rank representations~\cite{Li_2019_ICCV}.
The latter is inspired by the dynamic routing approaches such as~\cite{Guo_2019_CVPR} and channel pruning methods such as~\cite{Gordon_2018_CVPR,Liu_2017_ICCV}.

\subsection{Task-specific weight transform}
\label{sec:approach:weight_transform}
\begin{figure}[t]
    \centering
    \includegraphics[width=0.62\linewidth]{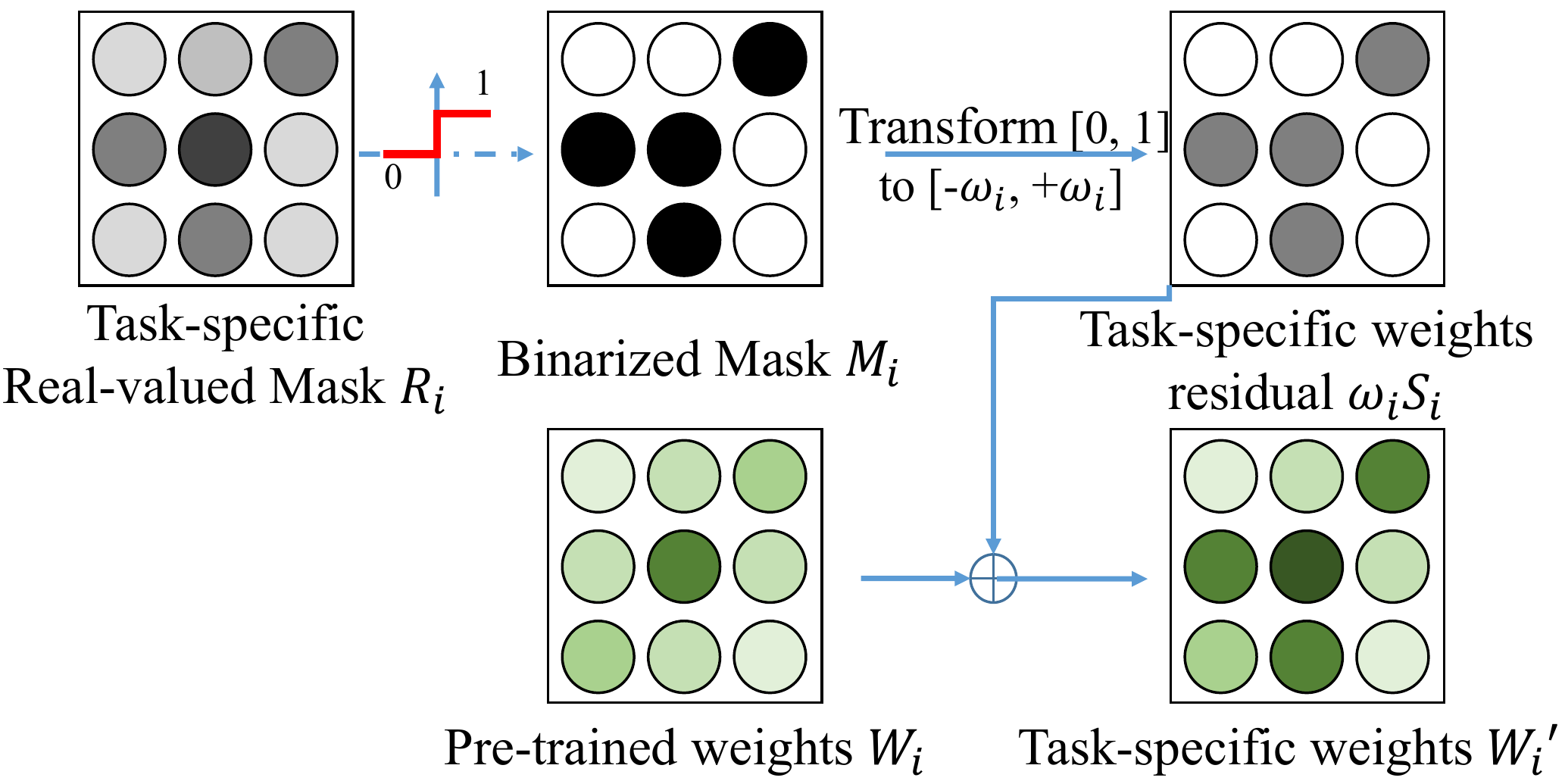}
    \caption{\textbf{The weight patching model.} The pre-trained weights 
    are augmented using a task-specific scaled sign matrix. For deployment, only the binary masks $\textbf{M}_i$ and the scaling factors $\omega_i$ are stored in the model patch.}
    \label{fig:affine}
\end{figure}
We denote the pre-trained weights by $\bm{\theta} = \{\mathbf{W}_1, \dots, \mathbf{W}_N\}$ and define the task-specific weight trasformation as:
\begin{equation} \label{eq:affine}
    \mathbf{W}_i'=\mathbf{W}_i + \omega_i\mathbf{S}_i
\end{equation}
where $\mathbf{S}_i$ is a 1-bit matrix of the same shape as $\mathbf{W}_i$ containing $\{-1, +1\}$ and $\omega_i$  is a scaling factor.
For implementation convenience we use $\mathbf{M}_i$ as a 1-bit mask tensor with values in $\{0, 1\}$, and define $\mathbf{S}_i=\bm{1} - 2 \mathbf{M}_i$. This formulation is equivalent to Eq.~2 in~\cite{Mancini_2018_ECCVW}, with $k_0$ set to 1, and $k_2$ set to $-2k_1$. The reason we instead chose our formulation is that we empirically found the learned $k_2$ in~\cite{Mancini_2018_ECCVW} to be roughly distributed as $-2k_1$. We thus directly formulated Eq.~\ref{eq:affine} as learning a properly scaled zero-centered residual.

The incremental footprint of the model in Eq.~\ref{eq:affine} is $\{\omega_i, \textbf{M}_i|~i\in \{1,\dots,N\}\}$, or roughly 1-bit per model weight, with a negligible additional cost for the per-layer scalar $\omega_i$.
To learn the 1-bit mask $\mathbf{M}_i$, we follow the same approach as~\cite{Mallya_2018_ECCV} and~\cite{Mancini_2018_ECCVW}. We define the process in Eq.~\ref{eq:binarize_mask}, and illustrate it in Fig.~\ref{fig:affine}.
During training, real-valued mask variables are maintained ($\mathbf{R}_i$ in Eq.~\ref{eq:binarize_mask}) and differentiable binarization is applied to learn the masks. The binarization function is a hard thresholding function, and its gradient is set to that of the \textit{sigmoid} function $\sigma$. After training, only the binarized masks and the per-layer scaling factors are used to deploy the model patch.
\begin{equation} \label{eq:binarize_mask}
    \mathbf{M}_i=\binarize(\mathbf{R}_i)
\end{equation}

\subsection{Spot patching}

Assuming a 32-bit float representation, the task-specific weight transform produces patches 1/32 the size of the original network, regardless of the target dataset. However, our intuition is that the difficulty of adapting a model should also depend on the target dataset, so the footprint should vary for different tasks.

We design a gating mechanism to adapt the model complexity to different tasks. 
The process is defined in Eq.~\ref{eq:gating}, and Fig.~\ref{fig:gating} shows the idea. Simply speaking, we add a \textit{gate} $g_i$ for each network layer. The layer uses the original pre-trained weights if the gate value is 0; otherwise, the layer uses weight transform to update the parameters.
The benefit of using the gating indicator $g_i$ is that it allows to search for a task-specific subset of layers to patch, \textit{rather than patching all layers}. 
Compared to patching the whole network, it reduces the patch footprint to $\bm{\gamma}=\{\omega_i, \mathbf{M}_i|~i\in\{1,\dots,N\} ~ \text{and} ~ g_i=1\}$.
\begin{equation}
\label{eq:gating}
    \mathbf{W}_i'=\mathbf{W}_i + \underbrace{g_i}_{\text{gating}} \omega_i(\bm{1} - 2 \mathbf{M}_i)
\end{equation}
\begin{equation} \label{eq:binarize_gate}
    g_i=\binarize (f_i)
\end{equation}
To learn $g_i$, we simply use the same differentiable binarization trick as for learning the $\mathbf{M}_i$. In Eq.~\ref{eq:binarize_gate}, $f_i$ is real-valued, and it is only used during training.
To force the number of patched layers to be small, we minimize the number of patched layers $\sum_{i=1}^{N} g_i$ in the training loss (see next Section~\ref{sec:approach:final_model}).

SpotPatch gating module design follows the same vein of dynamic routing approaches, especially~\cite{Guo_2019_CVPR}. The difference with respect to~\cite{Guo_2019_CVPR} lies in the fact that SpotPatch applies binary quantization to the tuning route (Sec.~\ref{sec:approach:weight_transform}), greatly reducing the footprint. On the one hand, SpotPatch gating module can be seen as a simpler version of~\cite{Guo_2019_CVPR}, in that we use the same differentiable binarization trick for both generating the binary masks, and directly optimizing the gating variables. On the other hand, our loss function explicitly minimizes the number of patched layers (see next Section~\ref{sec:approach:final_model}), hence delivering a task-adaptive footprint, rather than a fix-sized one as in~\cite{Guo_2019_CVPR}. For example, \cite{Guo_2019_CVPR} would provide a~300\%, or~1000\% footprint increase (they provided two models) to solve 10 classification tasks, while our method only requires an extra~35\% footprint to solve 9 additional object detection tasks.

\subsection{Our final task-adaptive detector}
\label{sec:approach:final_model}

Our final model on a new task is similar to Eq.~\ref{eq:deep_net}, with the parameters replaced by Eq.~\ref{eq:gating}. During training, we use floating-point numbers and differentiable binarization (Eq.~\ref{eq:binarize_mask} and Eq.~\ref{eq:binarize_gate}). During deployment, bit representations are used to efficiently encode the learned patch.
Since the Batch Normalization layers did not constitute much of the footprint, we also trained task-specific Batch Normalization layers in addition to the convolutional weight residuals (\cite{Mancini_2018_ECCVW} also patches BN layers).
We use Eq.~\ref{eq:final_loss} to optimize the task-specific patch $\bm{\gamma}$:
\begin{equation} \label{eq:final_loss}
    L(\bm{\gamma})=L_{det}(\bm{\gamma}) + \lambda_{sps} L_{sps}(\bm{\gamma}) + \lambda_{adp} L_{adp}(\bm{\gamma})
\end{equation}
where: $L_{det}(\bm{\gamma})$ is the detection loss optimizing both the class confidence scores and the box locations. $L_{sps}(\bm{\gamma})=\sum_{i=1}^N g_i$ is the sparsity-inducing loss, pushing the number of patched layers to be small.
Finally, $L_{adp}(\bm{\gamma})=\sum_{i=1}^N \lVert\omega_i\rVert_2^2$ is the domain-adaptation loss forcing the scaling factors $\omega_i$ to be small, and thus $\bm{\theta}'$ to be similar to $\bm{\theta}$. 
In this way, the pre-trained general-purpose source model serves as a strong prior for the task-specific target model.
A similar loss to $L_{adp}$ has been employed in prior domain adaptation works~\cite{yang2007adapting,kovashka2013attribute}, to force the adapted weights to be small and train accurate models with limited data.
We provide an ablation study for $\lambda_{sps}$ in Sec.~\ref{sec:results:sparse}, while for $\lambda_{adp}$ we use the constant value of 2E-5, as selected in preliminary experiments.
\section{Experiments}
\label{sec:results}
We propose two scenarios for patching a mobile object detector, and design experiments to validate our model for both use-cases.

\textit{Adapting an object detector to solve a new task.}
For this scenario (Sec.~\ref{sec:results:detection_decathlon}), assume that we released a mature generic object detector to our users. However, if the user wants to perform a new unsupported detection task, we need to adapt the generic detector to solve the new task.
For example, we may transform it into a product detector, or we may turn it into a pet detector.
In this scenario, the challenge is to accurately and efficiently solve the new task.

\textit{Updating an existing detection model, whenever additional training data for the same task is available.}
For this use-case (Sec.~\ref{sec:results:model_updating}), suppose we released an initial mobile model. We gather more training data after the model is released to the users.
We then want to update the users models to improve the accuracy, while keeping the download byte size to be small.
In this case, we assume that there is no significant shift in the data distribution, yet the initial model may be inaccurate because of the cold start.

In addition to the above settings, in Sec.~\ref{sec:results:8bit_models} we consider the more practical 8-bit model quantization scenario. In Sec.~\ref{sec:results:sparse} we study the effect of the sparsity constraint. In Sec.~\ref{sec:results:visualization} we provide visualizations of the learned model patches.

\paragraph{Implementation details.}
Our experimental configuration is based on the SSD-FPNLite architecture~\cite{ssd_fpnlite,Lin_2017_ICCV}, a practical mobile-friendly architecture.
Slightly departing from the original configuration, we use the MobileNetV2~\cite{Sandler_2018_CVPR} architecture with $320\times320$ inputs.

\paragraph{Baselines.}
We compare with the following transfer learning baselines:

\begin{itemize}[nolistsep,noitemsep]
    \item \textsc{Fine-Tuning}~\cite{Huang_2017_CVPR}: This method fine-tunes the whole network. It provides a strong baseline in terms of accuracy, at the expense of a very large footprint.
    
    \item \textsc{Tower Patch}~\cite{Huang_2017_CVPR}: This method re-trains only the parameters in the detection head of the model. It is an adaptation and enhancement of the classifier last layer fine-tuning method, for object detection.
    
    \item \textsc{Bn Patch}~\cite{Mudrakarta_2019_ICLR}, \textsc{Dw Patch}~\cite{Mudrakarta_2019_ICLR}, \textsc{Bn+Dw Patch}~\cite{Guo2019DepthwiseCI,Mudrakarta_2019_ICLR}: These methods learn task-specific BatchNorm, Depthwise, or BatchNorm + Depthwise layers, respectively. They provide a patch with a tiny footprint.
    
    \item \textsc{Piggyback}~\cite{Mallya_2018_ECCV}: Learns task-specific binary masks, and uses element-wise multiplication to apply the masks and obtain the task-specific convolutional kernels. Since the masks are binary, the 1-bit patch has a very low footprint. 
    
    \item \textsc{WeightTrans}~\cite{Mancini_2018_ECCVW}: This baseline also relies on binary masks. It applies affine transformations on the masks to get the task-specific kernels.
\end{itemize}

We reproduced all approaches using the SSD-FPNLite architecture for the following reasons: 1) most papers only report results for classification tasks, or for a few detection datasets; 2) implementations based on different network architectures make comparing the footprint challenging. In all our experiments we thus use our implementation of these methods.
We did not re-implement and evaluate~\cite{Guo_2019_CVPR}, as this approach creates weight residuals with the same bit-size as the original ones, i.e. full float kernels. It is thus not effective at significantly reducing the patch footprint, requiring as large as 3x or 10x (compared to the original model) additional footprints on the Visual Decathlon Challenge.
As explained in next Section~\ref{sec:results:detection_decathlon}, we use finetuning as the baseline for parameter-\textit{inefficient} transfer learning.

\textit{Metrics.} To evaluate the detection performance, we use the mAP@0.5, which is the mean Average Precision over all classes, considering the generous threshold of $\text{IoU}\ge 0.5$. For the footprint, we report the ratio~ $\frac{\text{bitsize}(\bm{\gamma})}{\text{bitsize}(\bm{\theta})}$ between the size of the additional parameters $\bm{\gamma}$ necessary to solve the new tasks, and the size of the original model $\bm{\theta}$. Sec.~\ref{sec:results:detection_decathlon} and \ref{sec:results:model_updating} consider 32-bit float models, in which the footprint metric is identical to that of Visual Decathlon Challenge \cite{rebuffi2017learning}. In this case (32-bit float), the 1-bit binary masks (e.g., \cite{Mallya_2018_ECCV,Mancini_2018_ECCVW})
reduce the representation footprint by 32x. Sec.~\ref{sec:results:8bit_models} considers instead the 8-bit representation, as it is more relevant for mobile applications.

\subsection{Detection Decathlon}
\label{sec:results:detection_decathlon}

We use the OpenImages~V4~\cite{kuznetsova2018open} as the dataset for training the generic object detection model. This is a large-scale dataset featuring 1.74 million images of common objects from 600 different classes.
Our fully-trained mobile model achieves 27.6\% mAP@0.5 on the OpenImage~V4 validation set.
We then consider adapting the OpenImage~V4 pre-trained model to nine additional detection tasks (see Tab.~\ref{tab:datasets}), and compare models on the basis of how well they solve all the problems (mAP@0.5), and how small is their footprint.

\begin{table}[t]
    \tablefontsize
    \centering
    \caption{\textbf{Detection Decathlon datasets}. Number of samples and classes for each dataset used in our benchmark. Some datasets did not provide testing annotations; we thus evaluate on the held-out validation sets.}
    \setlength{\tabcolsep}{4pt}
    \begin{tabularx}{1.0\linewidth}{p{1.4cm}|*{2}{>{\centering\arraybackslash}X}|*{1}{>{\centering\arraybackslash}X}|| p{1.4cm}|*{2}{>{\centering\arraybackslash}X}|*{1}{>{\centering\arraybackslash}X}}
    \Xhline{2\arrayrulewidth}
        Name & \#Trainval & \#Eval & \#Classes &
        Name & \#Trainval & \#Eval & \#Classes \\
    \hline
        OID\cite{kuznetsova2018open} & 1,668,276 & - & 601 &
        Face\cite{Yang_2016_CVPR} & 12,880 & 3,226 & 1 \\
        Birds\cite{Welinder_2010_Caltech} & 3,000 & 3,033 & 200 &
        Kitti\cite{Geiger_2012_CVPR} & 6,981 & 500 & 2 \\
        Cars\cite{Krause_2013_ICCVW} & 8,144 & 8,041 & 196 &
        Pet\cite{Parkhi_2012_CVPR} & 3,180 & 500 & 37 \\
        COCO\cite{lin2014microsoft} & 118,287 & 5,000 & 80 &
        RPC\cite{Wei_2019_RPC} & 5,400 & 600 & 200 \\
        Dogs\cite{Khosla_2011_FGVC} & 12,000 & 8,580 & 120 &
        VOC\cite{Everingham10} & 16,551 & 4,952 & 20 \\
    \Xhline{2\arrayrulewidth}
    \end{tabularx}
    \label{tab:datasets}
\end{table}

For the datasets used in the Detection Decathlon,
the Caltech-UCSD Birds~\cite{Welinder_2010_Caltech} (Bird), Cars~\cite{Krause_2013_ICCVW} (Car), and Stanford Dogs~\cite{Khosla_2011_FGVC} (Dog) are fine-grained categorization datasets. They provide center-view objects with bounding box annotations.
The WiderFace~\cite{Yang_2016_CVPR} (Face) and Kitti~\cite{Geiger_2012_CVPR} (Kitti) are human-related datasets. The former features human faces in different contexts and scales; the latter features vehicles and pedestrians for self-driving studies.
The Oxford-IIIT Pet~\cite{Parkhi_2012_CVPR} (Pet) and Retail Product Checkout~\cite{Wei_2019_RPC} (RPC) require both the fine-grained classification as well as localization. They involve many categories that appear in different locations and scales.
Finally, the Pascal VOC~\cite{Everingham10} (VOC) and COCO~\cite{lin2014microsoft} (COCO) are common object detection datasets. The class labels defined in them are subsets of those in OpenImage~V4.

Different from the Visual Decathlon challenge, we assume the performance and footprint of the model on the original task (OpenImages~V4) are unchanged, and only compare the models accuracy and footprint on the remaining nine tasks. We refer to the above problem as the \textit{Detection Decathlon} problem.
Similarly to~\cite{rebuffi2017learning} we also provide a decathlon-inspired scoring function to evaluate the performance of each method on the benchmark:
\begin{equation}
    Score =  10000 \frac{1}{D}\sum_{d=1}^D \left(\frac{\left| s_d - b_d \right|^+}{1-b_d}\right)^2
\end{equation}
where: the score $s_d$ is the mAP of the considered approach on the $d$ task; $b_d$ is the score of a strong \textit{baseline} on the $d$ task; 10,000 is the maximum achievable score, and $D$ is the total number of tasks to be solved. Similarly to~\cite{rebuffi2017learning}, we select $b_d$ to be the mAP of \textsc{Fine-Tuning} on task $d$, and normalize it so that its total score on the benchmark is 2,500. Specifically we set: $b_d~=~2~\text{mAP}_d(\text{\textsc{Fine-Tuning}})~-~1$.

To compare efficiency and effectiveness of different methods using one single metric, we finally report the \textit{Score/Footprint} ratio~\cite{Mancini_2018_ECCVW}. For a given method, this metric practically measures the performance achieved for every Mb of footprint.


\begin{table*}[t]
    \tablefontsize
    \centering
    \caption{\textbf{Detection Decathlon}. Footprint, per-dataset mAP, Average mAP, Score and Score / Footprint for each method. Best method (other than fine-tuning) in \textbf{bold}, second-best \underline{underlined}.  High score, low footprint, and high score/footprint ratio is good. Ours is the most parameter-efficient method in terms of Score/Footprint. It achieves mAP and Score comparable to the most accurate approach (\textsc{WeightTrans 65.2\%}), with a 24\% reduction in footprint.}
    \setlength{\tabcolsep}{1pt}
    \begin{tabularx}{1.0\linewidth}{p{2.3cm}|*{1}{>{\centering\arraybackslash}X}|*{9}{>{\centering\arraybackslash}X}|*{1}{>{\centering\arraybackslash}X}|*{2}{>{\centering\arraybackslash}X}}
    \Xhline{2\arrayrulewidth}
        Method 
        & \rotatebox{90}{\parbox{1.3cm}{\centering \scriptsize Footprint}}
        & \rotatebox{90}{\parbox{1.3cm}{\centering Bird}}
        & \rotatebox{90}{\parbox{1.3cm}{\centering Car}}
        & \rotatebox{90}{\parbox{1.3cm}{\centering COCO}}
        & \rotatebox{90}{\parbox{1.3cm}{\centering Dog}}
        & \rotatebox{90}{\parbox{1.3cm}{\centering Face}} 
        & \rotatebox{90}{\parbox{1.3cm}{\centering Kitti}}
        & \rotatebox{90}{\parbox{1.3cm}{\centering Pet}}
        & \rotatebox{90}{\parbox{1.3cm}{\centering RPC}}
        & \rotatebox{90}{\parbox{1.3cm}{\centering VOC}}
        & \rotatebox{90}{\parbox{1.3cm}{\centering Average mAP}}
        & \rotatebox{90}{\parbox{1.3cm}{\centering Score}}
        & \rotatebox{90}{\parbox{1.3cm}{\centering Score / \\Footprint}} \\
    \hline
        \textsc{Fine-Tuning} & 9.00 & 40.8 & 90.4 & 39.7 & 68.5 & 35.6 & 71.9 & 90.9 & 99.5 & 68.3 & 67.3 & 2500 & 278\\
    \hline
        \textsc{Tower Patch} & 0.35 & 10.0 & 25.5 & 31.4 & 21.2 & 29.1 & 49.7 & 66.1 & 87.6 & 70.6 & 43.5 & 827 & 2362 \\
        \textsc{Bn Patch} & \textbf{0.19} & 22.6 & 71.6 & 30.2 & 47.8 & 26.0 & 50.7 & 80.6 & 92.0 & \textbf{71.1} & 54.7 & 910 & \underline{4789} \\
        \textsc{Dw Patch} & 0.34 & 22.6 & 69.2 & 30.7 & 43.6 & 26.4 & 52.1 & 80.0 & 92.6 & \underline{70.8} & 54.2 & 898 & 2642 \\
        \textsc{Bn+Dw Patch} & 0.50 & 27.3 & 80.9 & 31.0 & 52.7 & 28.0 & 53.1 & 83.3 & 95.7 & 70.6 & 58.1 & 1012 & 2023 \\
        \textsc{Piggyback} & \underline{0.30} & 32.2 & 87.5 & 32.4 & 60.8 & 28.6 & 57.4 & 87.7 & 97.0 & 66.0 & 61.1 & 1353 & 4509 \\
        \textsc{WeightTrans} & 0.46 & \textbf{36.6} & \textbf{90.3} & \textbf{37.2} & \textbf{66.6} & \textbf{30.6} & \textbf{65.3} & \textbf{90.5} & \underline{98.7} & 70.7 & \textbf{65.2} & \textbf{1987} & 4319 \\
    \hline
        \textsc{Ours} & 0.35 & \underline{35.8} & \underline{89.8} & \underline{36.6} & \underline{63.3} & \underline{30.1} & \underline{64.0} & \underline{90.3} & \textbf{98.9} & 70.6 & \underline{64.4} & \underline{1858} & \textbf{5310} \\
    \Xhline{2\arrayrulewidth}
    \end{tabularx}
    \label{tab:model_adaptation}
\end{table*}

\begin{figure*}[t]
    \centering
    \includegraphics[width=1.0\linewidth]{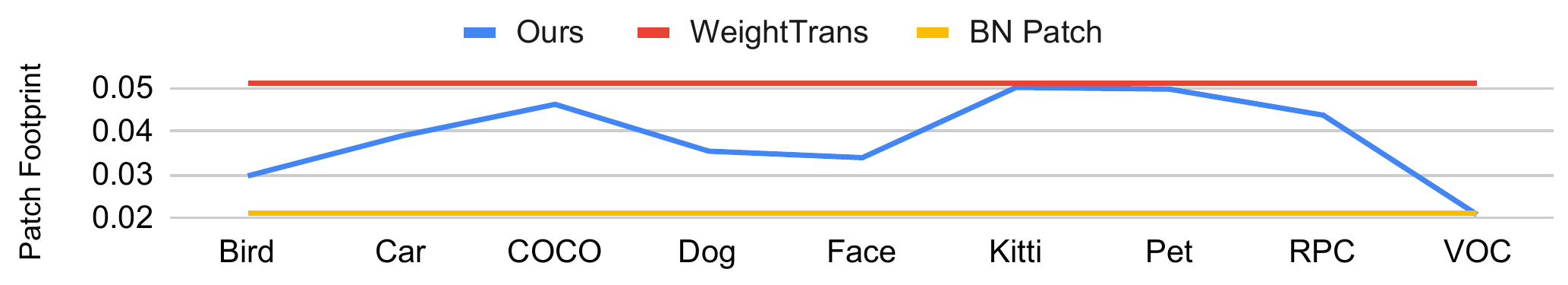}
    \caption{\textbf{Per-dataset footprint.} \textsc{SpotPatch} footprint varies between that of  \textsc{Piggyback} and that of \textsc{BN Patch}, depending on the complexity of the task.}
    \label{fig:per_dataset_patch_size}
\end{figure*}

\begin{table}[t]
    \tablefontsize
    \centering
    \caption{\textbf{Model updating}. Our method achieves comparable mAP and smaller footprint than the strongest baseline. The percentages (e.g. 10\%) indicate the amount of data to train the source model. The best method is shown in \textbf{bold}, second-best \underline{underlined}. Only one footprint number is shown for the baseline methods because they generate the same footprint regardless of the task.}
    \setlength{\tabcolsep}{1pt}
    \begin{tabularx}{0.75\linewidth}{p{2.3cm}|*{4}{>{\centering\arraybackslash}X}|*{4}{>{\centering\arraybackslash}X}}
    \Xhline{2\arrayrulewidth}
        \multirow{2}{*}{Method} & \multicolumn{4}{c|}{Footprint (\%)} & \multicolumn{4}{c}{mAP (\%)} \\
        & 10\% & 20\% & 40\% & 80\% & 10\% & 20\% & 40\% & 80\% \\
    \hline
        \textsc{Fine-Tuning} & \multicolumn{4}{c|}{100.0} & 37.6 & 37.6 & 38.2 & 38.5 \\
    \hline
        \textsc{Tower Patch} & \multicolumn{4}{c|}{3.85} & 24.5 & 26.7 & 32.4 & 35.8 \\
        \textsc{Bn Patch} & \multicolumn{4}{c|}{\textbf{2.08}} & 26.2 & 28.1 & 33.0 & 35.8 \\ 
        \textsc{Dw Patch} & \multicolumn{4}{c|}{3.76} & 25.9 & 27.8 & 33.1 & \textbf{36.0} \\ 
        \textsc{Bn+Dw Patch} & \multicolumn{4}{c|}{5.59} & 26.8 & 28.7 & 33.4 & \underline{35.9} \\ 
        \textsc{Piggyback} & \multicolumn{4}{c|}{3.32} & 26.4 & 28.3 & 32.0 & 35.3 \\ 
        \textsc{WeightTrans} & \multicolumn{4}{c|}{5.15} & \textbf{32.7} & \textbf{32.4} & \textbf{34.8}  & \underline{35.9} \\ 
    \hline
        \textsc{Ours} & 5.04 & 4.98 & 4.21 & \textbf{2.08} & \underline{32.0} & \underline{31.4} & \underline{34.2} & 35.8 \\
    \Xhline{2\arrayrulewidth}
    \end{tabularx}
    \label{tab:model_updating}
\end{table}


Tab.~\ref{tab:model_adaptation} and  Fig.~\ref{fig:per_dataset_patch_size} shows our main results.
Our first observation is that \textit{patching an object detector is more challenging than patching a classifier.} Notably, none of the tested methods matches the \textsc{Fine-Tuning} Score, or mAP. 
The enhanced last layer fine-tuning method, \textsc{Tower Patch}, only achieves 33\% of the \textsc{Fine-Tuning} score.
Patching dispersed bottleneck layers provides reasonable improvements. For example in terms of Score, \textsc{Bn Patch}, \textsc{DW Patch}, and \textsc{Bn+Dw Patch} are 10.0\%, 8.6\%, and 22.4\% relatively better than \textsc{Tower Patch}, and they all provide less than 0.50 footprints.
However, the gap with respect to \textsc{Fine-Tuning} is still large. They only maintain less than 40.5\% of the \textsc{Fine-Tuning} Score.
The kernel quantization methods \textsc{Piggyback} and \textsc{WeightTrans} maintain at least 54.1\% of the \textsc{Fine-Tuning} Score, while keeping the footprint below 0.46. 
\textsc{SpotPatch} achieves comparable performance to \textsc{WeightTrans} at only a 0.35 footprint. It also maintains 74.3\% of the \textsc{Fine-Tuning} Score.

\textit{Our approach provides the best tradeoff by being parameter-efficient and yet accurate, as measured by the Score/Footprint ratio. This is achieved by learning patches with a task-adaptive footprint, resulting on average in a 24\% footprint reduction with respect to \textsc{WeightTrans}, with only minor loss in accuracy.}

\subsection{Model updating}
\label{sec:results:model_updating}

For this experiment, we use the COCO dataset~\cite{lin2014microsoft}.
First, we trained detection models (initialized from  ImageNet\cite{deng2009imagenet} pre-trained model) on 10\%, 20\%, 40\%, and 80\% of the COCO training set. These models achieved 20.9\%, 24.2\%, 31.4\%, and 35.7\% mAP@0.5 on the COCO17 validation set, respectively.
Then, we applied different patching approaches to update these imprecise models, using 100\% of the COCO data.
We then compared mAP@0.5 of the patched models, as well as the resulting patch footprints.

Tab.~\ref{tab:model_updating} shows the results. Similar to the Detection Decathlon, we observe that none of the tested approaches is able to achieve the same mAP as fine-tuning. Ours is the only method that can adapt the footprint according to the source model quality and the amount of new training data:
At 10\% training data, we achieve comparable mAP as \textsc{WeightTrans} (32.0\% v.s. 32.7\%) at a comparable footprint (5.04\% v.s. 5.15\%). However, when more data is available, the patch footprint generated by our approach is smaller than \textsc{WeightTrans} (2.08\% v.s. 5.15\%), while accuracy remains comparable (35.8\% v.s. 35.9\%).

\textit{To summarize, our method can effectively adapt the patch footprint to the amount of new data to be learned by the patch, while maintaining high accuracy.}

\subsection{Accuracy-footprint tradeoff in 8-bit models}
\label{sec:results:8bit_models}

\begin{figure}[t]
    \centering
    \begin{subfigure}[b]{0.48\textwidth}
        \centering
        \includegraphics[width=1.0\linewidth]{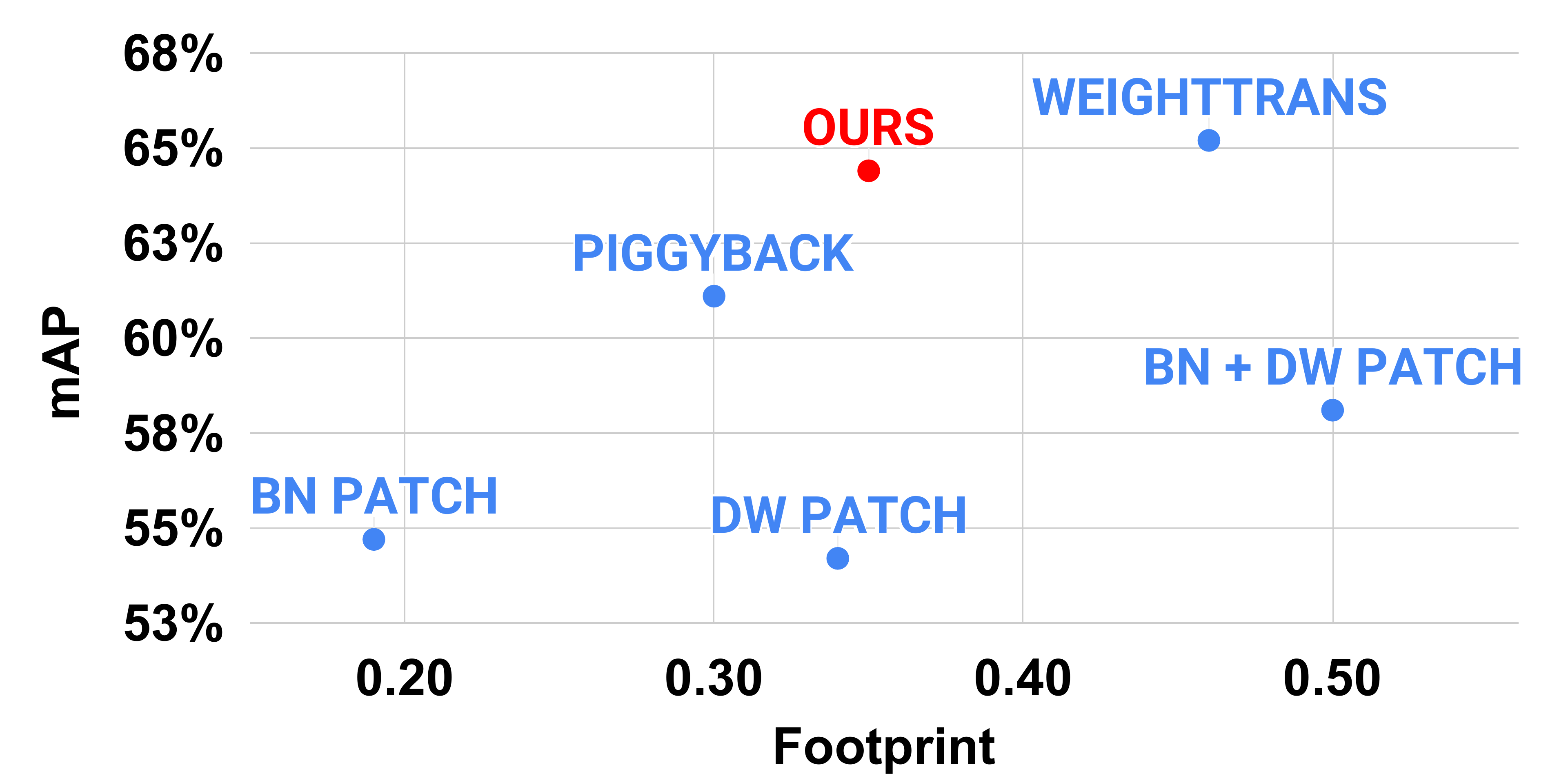}
        \caption{32-bit models}
    \end{subfigure}
    ~
    \begin{subfigure}[b]{0.48\textwidth}
        \centering
        \includegraphics[width=1.0\linewidth]{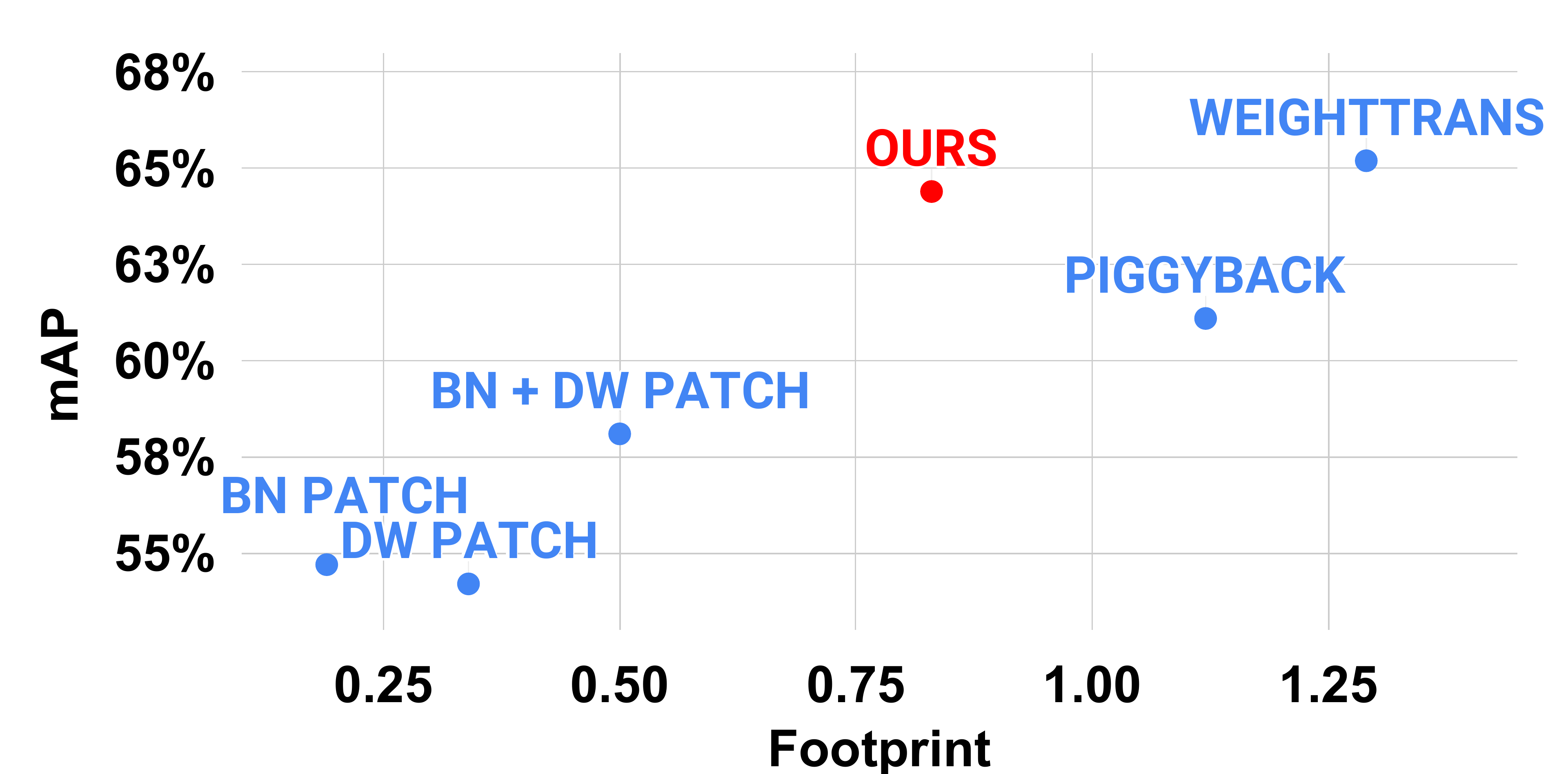}
        \caption{8-bit models}
    \end{subfigure}
    \caption{\textbf{Detection Decathlon: mAP v.s. Footprint.} We show the object detection mAP versus footprint, for both 32-bit and 8-bit models. The x-axis denotes the patch footprint for the Detection Decathlon while the y-axis indicates the detection performance measured by mAP@0.5. We expect a good model to have both a high mAP and a low footprint.}
    \label{fig:performance_and_average_precision}
\end{figure}

To compute the footprint, both Sec.~\ref{sec:results:detection_decathlon} and \ref{sec:results:model_updating} account for the binary mask size as 1/32 of the float kernel size. 
This convention is widely accepted by the participants in the Visual Decathlon Challenge.
However, in practical mobile applications, quantization-aware training~\cite{baskin2018uniq,Jacob_2018_CVPR,Hubara_2017_JMLR,quantization_aware_training} is often used to train a model using 8 bits per weight -- i.e., reduce the model size by 4x, without losing accuracy.
In the 8-bit model scenario, the relative footprint gains achieved by binary masks are thus 4x smaller than in the 32-bit model scenario.
We estimate the footprint of 8-bit models in the Detection Decathlon and assume quantization-aware training does not significantly hurt the detection performance~\cite{Jacob_2018_CVPR}. I.e., we did not train the 8-bit models but assume the mAP to be roughly the same as the 32-bit counterpart. To compare with the relative gains in the 32-bit scenarion, we show both of them side-by-side in Fig.~\ref{fig:performance_and_average_precision}.

As shown in Fig.~\ref{fig:performance_and_average_precision}, in the 8-bit scenario our method becomes more parameter-efficient than \textsc{Piggyback} and \textsc{WeightTrans}. The reason lies in the fact that ours is the only mask-based approach to explicitly minimize the number of masks in each patch. 
In Tab.~\ref{tab:model_size}, our model is thus as much as 26\% and 36\% more parameter-efficient than \textsc{Piggyback} and \textsc{WeightTrans}, respectively (0.83 v.s. 1.12, 1.29).
Our method would save additional 0.9Mb in network costs compared to \textsc{WeightTrans} \textit{per download}. Please note that while adding more tasks does not directly translate into mAP losses, footprint gains keep cumulating. In practical mobile application this effect would be amplified, as the same patch would need to be downloaded as many times as there are users. 
We thus argue that in practical mobile applications the 36\% footprint reduction achieved by our method over \textsc{WeightTrans}, with only a 0.8\% average mAP loss (64.4\% v.s. 65.2\%), constitutes a \textit{significant improvement} over \textsc{WeightTrans}.

\begin{table*}[t]
    \tablefontsize
    \centering
    \caption{\textbf{Detection Decathlon: mAP and patch size}. The patch size is measured by the expected download bytes to solve the additional nine tasks. It depends on both the model backbone and the method. Excluding the final layer in the classification head, the size of a single Mobilenet-v2 SSD-FPNLite detector is 7.99Mb for a 32-bit model, and 2.00Mb for an 8-bit model.
    }
    \setlength{\tabcolsep}{0pt}
    \begin{tabularx}{1.0\linewidth}{p{1.4cm}|*{1}{>{\centering\arraybackslash}X}|*{7}{>{\centering\arraybackslash}X}|*{1}{>{\centering\arraybackslash}X}}
    \Xhline{2\arrayrulewidth}
        & \textsc{Type} & \textsc{Fine-Tuning} & \textsc{Tower Patch} & \textsc{Bn Patch} & \textsc{Dw Patch} & \textsc{Bn+Dw Patch} & \textsc{Piggy back} & \textsc{Weight Trans} & \textsc{Ours} \\
    \hline
        mAP (\%) & - & 67.3 & 43.5 & 54.7 & 54.2 & 58.1 & 61.1 & 65.2 & 64.4 \\
    \hline
        \multirow{2}{*}{Footprint} 
        & 32-bit & \multirow{2}{*}{9.00} & \multirow{2}{*}{0.35} & \multirow{2}{*}{0.19} & \multirow{2}{*}{0.34} & \multirow{2}{*}{0.50} & 0.30 & 0.46 & 0.35 \\
        & 8-bit  & & & & & & 1.12 & 1.29 & 0.83 \\
    \hline
        \multirow{2}{*}{Patch size}
        & 32-bit & 71.9Mb & 2.77Mb & 1.50Mb & 2.71Mb & 4.02Mb & 2.39Mb & 3.70Mb & 2.79Mb \\
        & 8-bit  & 18.0Mb & 692Kb & 374Kb & 677Kb & 1.01Mb & 2.25Mb & 2.58Mb & 1.66Mb \\
    \Xhline{2\arrayrulewidth}
    \end{tabularx}
    \label{tab:model_size}
\end{table*}

\textit{To summarize, in practical 8-bit scenarios our method can potentially reduce \textsc{WeightTrans} footprint by 36\%, with only a negligible loss in performance. It is also the most parameter-efficient mask-based method.}

\subsection{Impact of the sparsity constraint}
\label{sec:results:sparse}

Next, we show the tradeoff between footprint and performance can further be selected by tuning $\lambda_{sps}$.
We perform a study on the Detection Decathlon tasks: we vary $\lambda_{sps}$ while keeping all other hyper-parameters the same.

\begin{table}[t]
    \tablefontsize
    \centering
    \caption{\textbf{Impact of sparsity constraint}. We show the percentage of patched layers, relative footprint, and mAP regarding models trained using different $\lambda_{sps}$.}
    \setlength{\tabcolsep}{4pt}
    \begin{tabularx}{1.0\linewidth}{p{1.6cm}|*{3}{>{\centering\arraybackslash}X}|*{3}{>{\centering\arraybackslash}X}|*{3}{>{\centering\arraybackslash}X}}
    \Xhline{2\arrayrulewidth}
        & \multicolumn{3}{c|}{Patched layers (\%)} & \multicolumn{3}{c|}{Footprint (\%)} & \multicolumn{3}{c}{mAP (\%)} \\
        & 1E-03 & 1E-04 & 1E-05 & 1E-03 & 1E-04 & 1E-05 & 1E-03 & 1E-04 & 1E-05 \\
    \hline
        Bird & 12.7 & 8.5 & 7.0 & 3.93 & 2.97 & 2.93 & 35.8 & 35.8 & 34.0 \\
        Car & 12.7 & 25.4 & 29.6 & 3.39 & 3.89 & 3.92 & 89.3 & 89.8 & 89.7 \\
        COCO & 19.7 & 42.3 & 53.5 & 4.12 & 4.62 & 4.76 & 35.5 & 36.6 & 36.8 \\
        Dog & 22.5 & 19.7 & 87.3 & 4.14 & 3.54 & 5.09 & 64.9 & 63.3 & 66.0 \\
        Face & 29.6 & 32.4 & 95.8 & 3.45 & 3.39 & 5.12 & 28.6 & 30.1 & 30.4 \\
        Kitti & 31.0 & 69.0 & 81.7 & 4.08 & 5.01 & 4.77 & 58.8 & 64.0 & 63.3 \\
        Pet & 22.5 & 53.5 & 52.1 & 4.24 & 4.97 & 4.32 & 90.9 & 90.3 & 90.5 \\
        RPC & 42.3 & 53.5 & 67.6 & 4.67 & 4.37 & 4.68 & 98.3 & 98.9 & 98.9 \\
        VOC & 0.0 & 0.0 & 2.8 & 2.08 & 2.08 & 2.54 & 70.5 & 70.6 & 69.9 \\
    \Xhline{2\arrayrulewidth}
        & \multicolumn{3}{c|}{Avg Patched layers (\%)} & \multicolumn{3}{c|}{Sum Footprint (\%)} & \multicolumn{3}{c}{Avg mAP (\%)} \\
    \hline
        & 21.4 & 33.8 & 53.1 & 34.1 & 34.8 & 38.1 & 63.6 & 64.4 & 64.4 \\
    \Xhline{2\arrayrulewidth}
    \end{tabularx}
    \label{tab:sparse}
\end{table}

Tab.~\ref{tab:sparse} shows the results. We observed that the $\lambda_{sps}$ has a direct impact on the percentage of patched layers and the patch footprint. 
In general, a large $\lambda_{sps}$ value forces the footprint to be small, while a small $\lambda_{sps}$ leads to a more accurate model.
If we only use a small value ($\lambda_{sps}$=1.00E-05), the method still patches the majority of the model layers (53.1\% in average), with a corresponding mAP of 64.4\%.
However, if we increase $\lambda_{sps}$ to 1.00E-03, the proportion of patched layers is significantly reduced (21.4\%) and mAP is only slightly reduced to 63.6\%.
We use $\lambda_{sps}$=1.00E-4 throughout the paper.

Tab.~\ref{tab:sparse} also highlights how the patching difficulty on different tasks varies. For example, the VOC target task is the most similar to the OpenImages source task. Our method learned that updating the batch normalization layers is enough. It thus degraded to \textsc{Bn Patch}, as almost none of the layers were patched.

\begin{figure*}[t]
    \centering
    \includegraphics[width=1.0\linewidth]{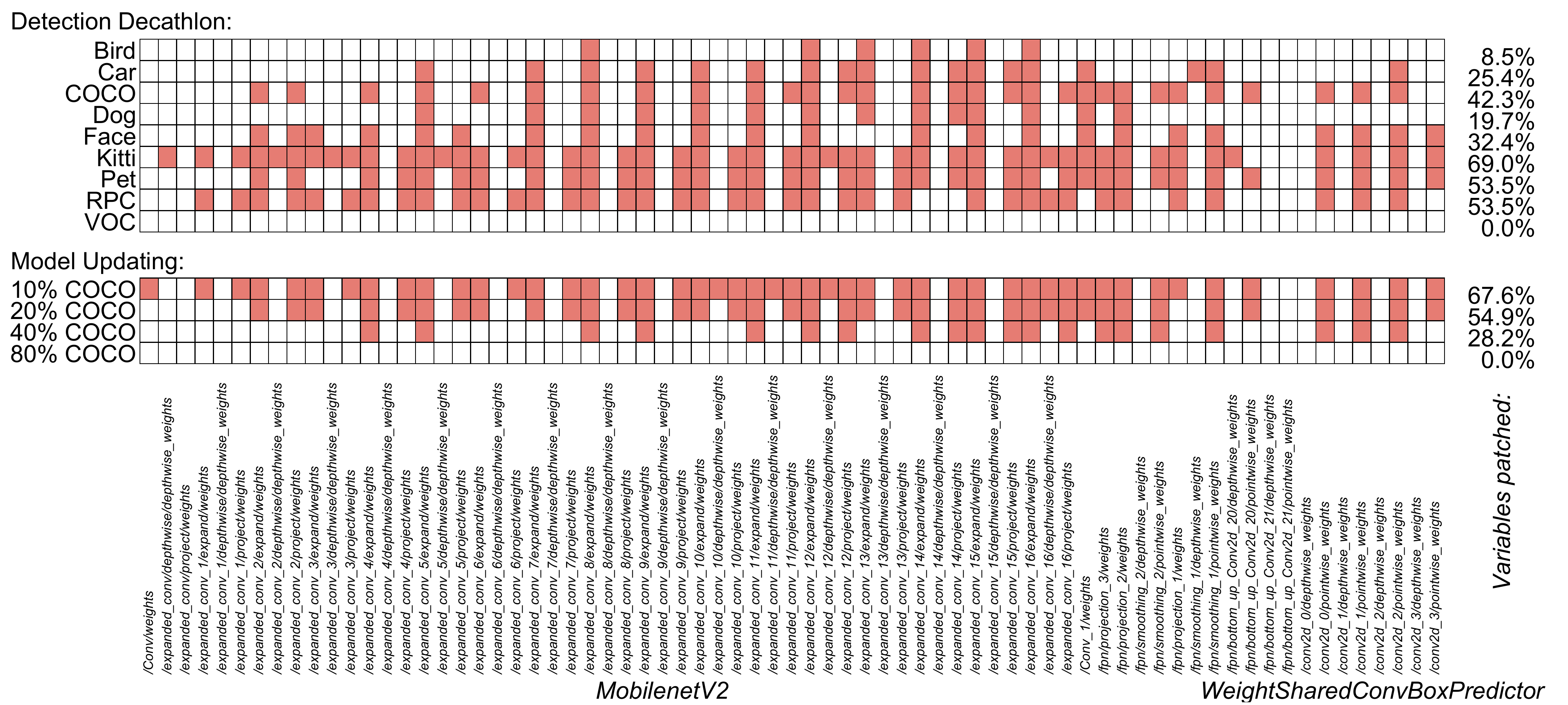}
    \caption{\textbf{Patched layers for different tasks.} We emphasize our method learned different model patches based on the complexity of the tasks. In the figure, each row denotes a patching policy for a specific dataset, and each column represents the gating indicator of a particular layer. For each row, a red block means that the learned model patch includes the weights residuals of the specific kernel. White blocks indicate that the source kernel is reused as is, with no contribution to the footprint. We show the proportion of patched layers on the right. Best viewed with 200\% zoom-in.}
    \label{fig:layers}
\end{figure*}

\subsection{Visualization of model patches}
\label{sec:results:visualization}
Next, we shed light on the nature of the patches learned by SpotPatch, and the effect of the source/target task similarity.
Fig.~\ref{fig:layers} shows the results for all the convolutional layers of the FPNLite model. 
For the Detection Decathlon problem, our approach patched fewer layers on the target tasks most similar to the source one, while modified more layers on the most dissimilar target tasks.
Our model learned that it is okay to leave all of the convolutional layers unchanged, for the VOC dataset. In this case, it degraded to the \textsc{Bn Patch} approach, which tunes only the batch normalization layers. The reason, we argue, is that VOC labels are a subset of the OpenImages~V4 labels.
In contrast, our model patched 69.0\% of the model layers for the Kitti task.
Though the Kitti dataset features everyday objects such as vehicles and pedestrians, the appearance of these objects is significantly different from OpenImages~V4 because they are captured by cameras mounted on cars for autonomous driving research.

Similar observations are made on the model updating. To patch the pre-trained models armed with 40\% and 80\% COCO information, our method patched 28.2\% and 0.0\% of the layers. However, for the imprecise and low-quality pre-trained models, for example, the models with 10\% and 20\% COCO information, our model patched 67.6\% and 54.9\% of the layers.

The patched models also shared some common patterns. Our method did not patch the first few convolutional layers and the FPN upsampling layers for most tasks.
We argue the reason is that these layers are responsible for recognizing fundamental visual features that can be shared across domains.
\section{Conclusion}
\label{sec:conclusion}
In this paper we drew the foundations for investigating parameter-efficient transfer learning in the context of mobile object detection. We introduced the Detection Decathlon problem, and provided the first systematic study of parameter-efficient transfer learning on this task. We proposed the SpotPatch approach, using task-specific weight transformations and dynamic routing to minimize the footprint of the learned patch. We also demonstrated how to use our technique for updating a pre-trained model. In all the considered benchmarks SpotPatch was shown to provide similar mAP as standard Weight-Transform, while being significantly more parameter-efficient. Additional potential gains were shown in the case of 8-bit quantization. We also noted how differently from classification benchmarks, none of the tested approaches was actually able to beat fine-tuning mAP, which calls for more work on the Detection Decathlon task.


\bibliographystyle{splncs}
\bibliography{egbib}

\end{document}